# The Complexity of Decentralized Control of Markov Decision Processes

**Daniel S. Bernstein, Shlomo Zilberstein,** and **Neil Immerman**
Department of Computer Science
University of Massachusetts
Amherst, Massachusetts 01003
{bern,shlomo,immerman}@cs.umass.edu

## Abstract

Planning for distributed agents with partial state information is considered from a decision-theoretic perspective. We describe generalizations of both the MDP and POMDP models that allow for decentralized control. For even a small number of agents, the finite-horizon problems corresponding to both of our models are complete for nondeterministic exponential time. These complexity results illustrate a fundamental difference between centralized and decentralized control of Markov processes. In contrast to the MDP and POMDP problems, the problems we consider *provably* do not admit polynomial-time algorithms and most likely require doubly exponential time to solve in the worst case. We have thus provided mathematical evidence corresponding to the intuition that decentralized planning problems cannot easily be reduced to centralized problems and solved exactly using established techniques.

## 1 Introduction

Among researchers in artificial intelligence, there has been growing interest in problems with multiple distributed agents working to achieve a common goal (Grosz & Kraus, 1996; Lesser, 1998; desJardins et al., 1999; Durfee, 1999; Stone & Veloso, 1999). In many of these problems, interagent communication is costly or impossible. For instance, consider two robots cooperating to push a box (Mataric, 1998). Communication between the robots may take time that could otherwise be spent performing physical actions. Thus, it may be suboptimal for the robots to communicate frequently. A planner is faced with the difficult task of deciding what each robot should do in between communications, when it only has access to its own sensory information. Other problems of planning for distributed agents with limited communication include maximizing the throughput of a multiple access broadcast channel (Ooi & Wornell, 1996) and coordinating multiple spacecraft on a mission together (Estlin et al., 1999). We are interested in the question of whether these planning problems are computationally harder to solve than problems that involve planning for a single agent or multiple agents with access to the exact same information.

We focus on centralized planning for distributed agents, with the Markov decision process (MDP) framework as the basis for our model of agents interacting with an environment. A partially observable Markov decision process (POMDP) is a generalization of an MDP in which an agent must base its decisions on incomplete information about the state of the environment (White, 1993). We extend the POMDP model to allow for multiple distributed agents to each receive local observations and base their decisions on these observations. The state transitions and expected rewards depend on the actions of all of the agents. We call this a decentralized partially observable Markov decision process (DEC-POMDP). An interesting special case of a DEC-POMDP satisfies the assumption that at any time step the state is uniquely determined from the current set of observations of the agents. This is denoted a decentralized Markov decision process (DEC-MDP). The MDP, POMDP, and DEC-MDP can all be viewed as special cases of the DEC-POMDP. The relationships among the models are shown in Figure 1.

There has been some related work in AI. Boutilier (1999) studies multi-agent Markov decision processes (MMDPs), but in this model, the agents all have access to the same information. In the framework we describe, this assumption is not made. Peshkin et al. (2000) use essentially the DEC-POMDP model (although they refer to it as a partially observable identical payoff stochastic game (POIPSG)) and discuss algorithms for obtaining approximate solutions to the corresponding optimization problem. The models that we study also exist in the control theory literature (Ooi et al., 1997; Aicardi et al., 1987). However, the computational complexity inherent in these models has not been studied. One closely related piece of work is that of Tsit-siklis and Athans (1985), in which the complexity of non-sequential decentralized decision problems is studied.

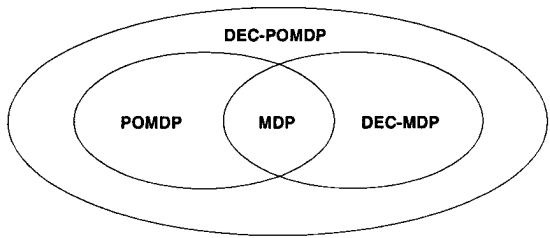

Figure 1: The relationships among the models.

We discuss the computational complexity of finding optimal policies for the finite-horizon versions of these problems. It is known that solving an MDP is P-complete and that solving a POMDP is PSPACE-complete (Papadimitriou & Tsitsiklis, 1987). We show that solving a DEC-POMDP with a constant number, $m \geq 2$, of agents is complete for the complexity class nondeterministic exponential time (NEXP). Furthermore, solving a DEC-MDP with a constant number, $m \geq 3$, of agents is NEXP-complete. This has a few consequences. One is that these problems *provably* do not admit polynomial-time algorithms. This trait is not shared by the MDP problems nor the POMDP problems. Another consequence is that any algorithm for solving either problem will most likely take doubly exponential time in the worst case. In contrast, the exact algorithms for finite-horizon POMDPs take "only" exponential time in the worst case. Thus, our results shed light on the fundamental differences between centralized and decentralized control of Markov decision processes. We now have mathematical evidence corresponding to the intuition that decentralized planning problems are more difficult to solve than their centralized counterparts. These results can steer researchers away from trying to find easy reductions from the decentralized problems to centralized ones and toward completely different approaches.

A precise categorization of the two-agent DEC-MDP problem presents an interesting mathematical challenge. The extent of our present knowledge is that the problem is PSPACE-hard and is contained in NEXP.

## 2 Centralized Models

A *Markov decision process (MDP)* models an agent acting in a stochastic environment to maximize its long-term reward. The type of MDP that we consider contains a finite set $S$ of states, with $s_0 \in S$ as the start state. For each state $s \in S$, $A_s$ is a finite set of actions available to the agent. $P$ is the table of transition probabilities, where $P(s'|s, a)$ is the probability of a transition to state $s'$ given that the agent performed action $a$ in state $s$. $R$ is the reward function, where $R(s, a)$ is the expected reward received by the agent given that it chose action $a$ in state $s$.

There are several different ways to define "long-term reward" and thus several different measures of optimality. In this paper, we focus on *finite-horizon* optimality, for which the aim is to maximize the expected sum of rewards received over $T$ time steps. Formally, the agent should maximize

$$E\left[\sum_{t=0}^{T} r(s_t, a_t)\right],$$

where $r(s_t, a_t)$ is the reward received at time step $t$. A *policy* $\delta$ for a finite-horizon MDP is a mapping from each state $s$ and time $t$ to an action $\delta(s, t)$. This is called a *non-stationary* policy. The decision problem corresponding to a finite-horizon MDP is as follows: Given an MDP $M$, a positive integer $T$, and an integer $K$, is there a policy that yields total reward at least $K$?

An MDP can be generalized so that the agent does not necessarily observe the exact state of the environment at each time step. This is called a *partially observable Markov decision process (POMDP)*. A POMDP has a state set $S$, a start state $s_0 \in S$, a table of transition probabilities $P$, and a reward function $R$, just as an MDP does. Additionally, it contains a finite set $\Omega$ of observations, and a table $O$ of observation probabilities, where $O(o|a, s')$ is the probability that $o$ is observed, given that action $a$ was taken and led to state $s'$. For each observation $o \in \Omega$, $A_o$ is a finite set of actions available to the agent. A policy $\delta$ is now a mapping from histories of observations $o_1, \ldots, o_t$ to actions in $A_{o_t}$. The decision problem for a POMDP is stated in exactly the same way as for an MDP.

## 3 Decentralized Models

A *decentralized partially observable Markov decision process (DEC-POMDP)* is a generalization of a POMDP to allow for distributed control by $m$ agents that may not be able to observe the exact state. A DEC-POMDP contains a finite set $S$ of states, with $s_0 \in S$ as the start state. The transition probabilities $P(s'|s, a^1, \ldots, a^m)$ and expected rewards $R(s, a^1, \ldots, a^m)$ depend on the actions of all agents. $\Omega^i$ is a finite set of observations for agent $i$, and $O$ is a table of observation probabilities, where $O(o^1, \ldots, o^m|a^1, \ldots, a^m, s')$ is the probability that $o^1, \ldots, o^m$ are observed by agents $1, \ldots, m$ respectively, given that the action tuple $\langle a^1, \ldots, a^m \rangle$ was taken and led to state $s'$. Each agent $i$ has a set of actions $A_o^i$ for each observation $o^i \in \Omega^i$. Notice that this model reduces to a POMDP in the one-agent case.

For each $a^1, \ldots, a^m, s'$, let $\omega(a^1, \ldots, a^m, s')$ denote the set of observation tuples that have a nonzero chance of occurring given that the action tuple $\langle a^1, \ldots, a^m \rangle$ was taken and led to state $s'$. To form a *decentralized Markov decision process (DEC-MDP)*, we add the requirement that for each $a^1, \ldots, a^m, s'$, and each $\langle o^1, \ldots, o^m \rangle \in \omega(a^1, \ldots, a^m, s')$ the state is uniquely determined by $\langle o^1, \ldots, o^m \rangle$. In the one-agent case, this model is essentially the same as an MDP.

We define a *local policy*, $\delta^i$, to be a mapping from *local histories of observations* $o_1^i, \ldots, o_t^i$ to actions $a^i \in A_{o_t}^i$. A *joint policy*, $\delta = \langle \delta^1, \ldots, \delta^m \rangle$, is defined to be a tuple of local policies. We wish to find a joint policy that maximizes the total expected return over the finite horizon. The decision problem is stated as follows: Given a DEC-POMDP $M$, a positive integer $T$, and an integer $K$, is there a joint policy that yields total reward at least $K$? Let **DEC-POMDP**$_m$ and **DEC-MDP**$_m$ denote the decision problems for the $m$-agent DEC-POMDP and the $m$-agent DEC-MDP, respectively.

## 4 Complexity Results

It is necessary to consider only problems for which $T < |S|$. If we place no restrictions on $T$, then the upper bounds do not necessarily hold. Also, we assume that each of the elements of the tables for the transition probabilities and expected rewards can be represented with a constant number of bits. With these restrictions, it was shown in (Papadimitriou & Tsitsiklis, 1987) that the decision problem for an MDP is P-complete. In the same paper, the authors showed that the decision problem for a POMDP is PSPACE-complete and thus probably does not admit a polynomial-time algorithm. We prove that for all $m \geq 2$, **DEC-POMDP**$_m$ is NEXP-complete, and for all $m \geq 3$, **DEC-MDP**$_m$ is NEXP-complete, where NEXP = NTIME$(2^{n^c})$ (Papadimitriou, 1994). Since P $\neq$ NEXP, we can be certain that there does not exist a polynomial-time algorithm for either problem. Moreover, there probably is not even an *exponential*-time algorithm that solves either problem.

For our reduction, we use a problem called **TILING** (Papadimitriou, 1994), which is described as follows: We are given a set of square tile types $T = \{t_0, \ldots, t_k\}$, together with two relations $H, V \subseteq T \times T$ (the horizontal and vertical compatibility relations, respectively). We are also given an integer $n$ in binary. A *tiling* is a function $f : \{0, \ldots, n-1\} \times \{0, \ldots, n-1\} \to T$. A tiling $f$ is *consistent* if and only if (a) $f(0,0) = t_0$, and (b) for all $i, j$ $(f(i,j), f(i+1,j)) \in H$, and $(f(i,j), f(i,j+1)) \in V$. The decision problem is to tell, given $T$, $H$, $V$, and $n$, whether a consistent tiling exists. It is known that **TILING** is NEXP-complete.

**Theorem 1** *For all $m \geq 2$, **DEC-POMDP**$_m$ is NEXP-complete.*

**Proof.** First, we will show that the problem is in NEXP. We can guess a joint policy $\delta$ and write it down in exponential time. This is because a joint policy consists of $m$ mappings from local histories to actions, and since $T < |S|$, all histories have length less than $|S|$. A DEC-POMDP together with a joint policy can be viewed as a POMDP together with a policy, where the observations in the POMDP correspond to the observation tuples in the DEC-POMDP. In exponential time, each of the exponentially many possible sequences of observations can be converted into belief states. The transition probabilities and expected rewards for the corresponding "belief MDP" can be computed in exponential time (Kaelbling et al., 1998). It is possible to use dynamic programming to determine whether the policy yields expected reward at least $K$ in this belief MDP. This takes at most exponential time.

Now we show that the problem is NEXP-hard. For simplicity, we consider only the two-agent case. Clearly, the problem with more agents can be no easier. We are given an arbitrary instance of **TILING**. From it, we construct a DEC-POMDP such that the existence of a joint policy that yields a reward of at least zero is equivalent to the existence of a consistent tiling in the original problem. Furthermore, $T < |S|$ in the DEC-POMDP that is constructed. Intuitively, a local policy in our DEC-POMDP corresponds to a mapping from tile positions to tile types, i.e., a tiling, and thus a joint policy corresponds to a pair of tilings. The process works as follows: In the position choice phase, two tile positions are randomly "chosen" by the environment. Then, at the tile choice step, each agent sees a different position and must use its policy to determine a tile to be placed in that position. Based on information about where the two positions are in relation to each other, the environment checks whether the tile types placed in the two positions could be part of one consistent tiling. Only if the necessary conditions hold do the agents obtain a nonnegative reward. It turns out that the agents can obtain a nonnegative *expected* reward if and only if the conditions hold for *all* pairs of positions the environment can choose, i.e., there exists a consistent tiling.

We now present the construction in detail. During the position choice phase, each agent only has one action available to it, and a reward of zero is obtained at each step. The states and the transition probability matrix comprise the nontrivial aspect of this phase. Recall that this phase intuitively represents the choosing of two tile positions. First, let the two tile positions be denoted $(i_1, j_1)$ and $(i_2, j_2)$, where $0 \leq i_1, i_2, j_1, j_2 \leq n-1$. There are $4 \log n$ steps in this phase, and each step is devoted to the choosing of one bit of one of the numbers. (We assume that $n$ is a power of two. It is straightforward to modify the proof to deal with the more general case.) The order in which the bits are chosen is important, and it is as follows: The bits of $i_1$ and $i_2$ are chosen from least significant up to most significant, alternating between the two numbers at each step. Then $j_1$ and $j_2$ are chosen in the same way. As the bits of the numbers are being determined, information about the relationships between the numbers is being recorded in the state. How we express all of this as a Markov process is explained below.

Each state has six components, and each component represents a necessary piece of information about the two tile positions being chosen. We describe how each of the components changes with time. A time step in our process can be viewed as having two parts, which we refer to as the stochastic part and the deterministic part. During the stochastic part, the environment "flips a coin" to choose either the number 0 or the number 1, each with equal probability. After this choice is made, the change in each component of the state can be described by a deterministic finite automaton that takes as input a string of 0's and 1's (the environment's coin flips). The semantics of the components, along with their associated automata, are described below:

1) Bit Chosen in the Last Step
This component of the state says whether 0 or 1 was just chosen by the environment. The corresponding automaton consists of only two states.

2) Number of Bits Chosen So Far
This component simply counts up to $4 \log n$, in order to determine when the position choice phase should end. Its automaton consists of $4 \log n + 1$ states.

3) Equal Tile Positions
After the $4 \log n$ steps, this component tells us whether the two tile positions chosen are equal or not. For this automaton, along with the following three, we need to have a notion of an accept state. Consider the following regular expression:

$$(00 + 11)^*.$$

Note that the DFA corresponding to the above expression, on an input of length $4 \log n$, ends in an accept state if and only if $(i_1, j_1) = (i_2, j_2)$.

4) Upper Left Tile Position
This component is used to check whether the first tile position is the upper left corner of the grid. Its regular expression is as follows:

$$(0(0+1))^*.$$

The corresponding DFA, on an input of length $4 \log n$, ends in an accept state if and only if $(i_1, j_1) = (0, 0)$.

5) Horizontally Adjacent Tile Positions
This component is used to check whether the first tile position is directly to the left of the second one. Its regular expression is as follows:

$$(10)^*(01)(11+00)^* \underbrace{(11+00) \cdots (11+00)}_{\log n}.$$

The corresponding DFA, on an input of length $4 \log n$, ends in an accept state if and only if $(i_1 + 1, j_1) = (i_2, j_2)$.

6) Vertically Adjacent Tile Positions
This component is used to check whether the first tile position is directly above the second one. Its regular expression is as follows:

$$\underbrace{(11+00) \cdots (11+00)}_{\log n}(10)^*(01)(11+00)^*.$$

The corresponding DFA, on an input of length $4 \log n$, ends in an accept state if and only if $(i_1, j_1 + 1) = (i_2, j_2)$.

So far we have described the six automata that determine how each of the six components of the state evolve based on input (0 or 1) from the environment. We can take the cross product of these six automata to get a new automaton that is only polynomially bigger and describes how the entire state evolves based on the sequence of 0's and 1's chosen by the environment. This automaton, along with the environment's "coin flips," corresponds to a Markov process. The number of states of the process is polylogarithmic in $n$, and hence polynomial in the size of the **TILING** instance. The start state $s_0$ is a tuple of the start states of the six automata. The table of transition probabilities for this process can be constructed in time polylogarithmic in $n$.

We have described the states, actions, state transitions, and rewards for the position choice phase, and we now describe the observation function. In this DEC-POMDP, the observations are uniquely determined from the state. For the states after which a bit of $i_1$ or $j_1$ has been chosen, agent one observes the first component of the state, while agent two observes a dummy observation. The reverse is true for the states after which a bit of $i_2$ or $j_2$ has been chosen. Intuitively, agent one "sees" only $(i_1, j_1)$, and agent two "sees" only $(i_2, j_2)$.

When the second component of the state reaches its limit, the tile positions have been chosen, and the last four components of the state contain information about the tile positions and how they are related. Of course, the exact tile positions are not recorded in the state, as this would require exponentially many states. This marks the end of the position choice phase. In the next step, which we call the tile choice step, each agent has $k + 1$ actions available to it, corresponding to each of the tile types, $t_0, \ldots, t_k$. We denote agent one's choice $t^1$ and agent two's choice $t^2$. No matter which actions are chosen, the state transitions deterministically to some final state. The reward function for this step is the nontrivial part. After the actions are chosen, the following statements are checked for validity:

1) If $(i_1, j_1) = (i_2, j_2)$, then $t^1 = t^2$.
2) If $(i_1, j_1) = (0, 0)$, then $t^1 = t_0$.
3) If $(i_1 + 1, j_1) = (i_2, j_2)$, then $(t^1, t^2) \in H$.
4) If $(i_1, j_1 + 1) = (i_2, j_2)$, then $(t^1, t^2) \in V$.

If all of these are true, then a reward of 0 is received. Otherwise, a reward of -1 is received. This reward function can be computed from the **TILING** instance in polynomial time. To complete the construction, the horizon $T$ is set to $4 \log n$ (exactly the number of steps it takes the process to reach the tile choice step, and fewer than the number of states $|S|$).

Now we argue that the expected reward is zero if and only if there exists a consistent tiling. First, suppose a consistent tiling exists. This tiling corresponds to a local policy for an agent. If each of the two agents follows this policy, then no matter which two positions are chosen by the environment, the agents choose tile types for those positions so that the conditions checked at the end evaluate to true. Thus, no matter what sequence of 0's and 1's the environment chooses, the agents receive a reward of zero. Hence, the expected reward for the agents is zero.

For the converse, suppose the expected reward is zero. Then the reward is zero no matter what sequence of 0's and 1's the environment chooses, i.e., no matter which two tile positions are chosen. This implies that the four conditions mentioned above are satisfied for any two tile positions that are chosen. The first condition ensures that for all pairs of tile positions, if the positions are equal, then the tile types chosen are the same. This implies that the two agents' tilings are exactly the same. The last three conditions ensure that this tiling is consistent. □

**Theorem 2** *For all $m \geq 3$, $\text{DEC-MDP}_m$ is NEXP-complete.*

**Proof.** (Sketch) Inclusion in NEXP follows from the fact that a DEC-MDP is a special case of a DEC-POMDP. For NEXP-hardness, we can reduce a DEC-POMDP with two agents to a DEC-MDP with three agents. We simply add a third agent to the DEC-POMDP and impose the following requirement: The state is uniquely determined by just the third agent's observation, but the third agent always has just one action and cannot affect the state transitions or rewards received. It is clear that the new problem qualifies as a DEC-MDP and is essentially the same as the original DEC-POMDP. □

The reduction described above can also be used to construct a two-agent DEC-MDP from a POMDP and hence show that $\text{DEC-MDP}_2$ is PSPACE-hard. However, this technique is not powerful enough to prove the NEXP-hardness of the problem. In fact, the question of whether $\text{DEC-MDP}_2$ is NEXP-hard remains open. Note that in the reduction in the proof of Theorem 1, the observation function is such that there are some parts of the state that are hidden from *both* agents. This needs to somehow be avoided in order to reduce to a two-agent DEC-MDP. A simpler task may actually be to derive a better *upper* bound for the problem. For example, it may be possible that $\text{DEC-MDP}_2 \in \text{co-NEXP}$, where $\text{co-NEXP} = \{\overline{L} | L \in \text{NEXP}\}$. Regardless of the outcome, the problem provides an interesting mathematical challenge.

## 5 Discussion

Using the tools of worst-case complexity analysis, we analyzed two models of decision-theoretic planning for distributed agents. Specifically, we proved that the finite-horizon $m$-agent DEC-POMDP problem is NEXP-complete for $m \geq 2$ and the finite-horizon $m$-agent DEC-MDP problem is NEXP-complete for $m \geq 3$.

The results have some theoretical implications. First, unlike the MDP and POMDP problems, the problems we studied *provably* do not admit polynomial-time algorithms, since $P \neq \text{NEXP}$. Second, we have drawn a connection between work on Markov decision processes and the body of work in complexity theory that deals with the exponential jump in complexity due to decentralization (Peterson & Reif, 1979; Babai et al., 1991). Finally, the two-agent DEC-MDP case yields an interesting open problem. The solution of the problem may imply that the difference between planning for two agents and planning for more than two agents is a significant one in the case where the state is collectively observed by the agents.

There are also more direct implications for researchers trying to solve problems of planning for distributed agents. Consider the growing body of work on algorithms for obtaining exact or approximate solutions for POMDPs (e.g., Jaakkola et al., 1995; Cassandra et al., 1997; Hansen, 1998). It would have been beneficial to discover that a DEC-POMDP or DEC-MDP is just a POMDP "in disguise," in the sense that it can easily be converted to a POMDP and solved using established techniques. We have provided evidence to the contrary, however. The complexity results do not answer all of the questions surrounding how these problems should be attacked, but they do suggest that the fundamentally different structure of the decentralized problems may require fundamentally different algorithmic ideas.

Finally, consider the infinite-horizon versions of the aforementioned problems. It has recently been shown that the infinite-horizon POMDP problem is undecidable (Madani et al., 1999) under several different optimality criteria. Since a POMDP is a special case of a DEC-POMDP, the corresponding DEC-POMDP problems are also undecidable. In addition, because it is possible to reduce a POMDP to a two-agent DEC-MDP, the DEC-MDP problems are also undecidable.

### Acknowledgments

The authors thank Micah Adler, Andy Barto, Dexter Kozen, Victor Lesser, Frank McSherry, Ted Perkins, and Ping Xuan for helpful discussions. This work was supported in part by the National Science Foundation under grants IRI-9624992, IRI-9634938, and CCR-9877078 and an NSF Graduate Fellowship to Daniel Bernstein.